\documentclass{article}

\usepackage{microtype}
\usepackage{graphicx}
\usepackage{subfigure}
\usepackage{booktabs} % for professional tables
\usepackage{subcaption}

\usepackage{cancel}
\usepackage{hyperref}
\usepackage{multirow}
\usepackage{float}
\usepackage[accepted]{icml2025}

\usepackage{amsmath}
\usepackage{amssymb}
\usepackage{mathtools}
\usepackage{amsthm}

\usepackage[capitalize,noabbrev]{cleveref}

\theoremstyle{plain}

\theoremstyle{definition}

\theoremstyle{remark}

\usepackage[textsize=tiny]{todonotes}

\icmltitlerunning{Examining Two Hop Reasoning Through Information Content Scaling}

\begin{document}

\twocolumn[
\icmltitle{Examining Two Hop Reasoning Through Information Content Scaling}




\begin{icmlauthorlist}
\icmlauthor{David Johnston}{eai}
\icmlauthor{Sergio Sanz}{}
\icmlauthor{Nora Belrose}{eai}
\end{icmlauthorlist}

\icmlaffiliation{eai}{EleutherAI}

\icmlcorrespondingauthor{David Johnston}{david@eleuther.ai}

\icmlkeywords{Machine Learning, ICML}

\vskip 0.3in
]



\printAffiliationsAndNotice{}  % leave blank if no need to mention equal contribution

\begin{abstract}
Prior work has found that transformers have an inconsistent ability to learn to answer latent two-hop questions -- questions of the form ``Who is Bob's mother's boss?'' We study why this is the case by examining how transformers' capacity to learn two-hop questions and answers (two-hop QA) scales with their size, motivated by prior work on transformer knowledge capacity for simple fact memorization. We find that capacity scaling and generalization both support the hypothesis that latent two-hop QA requires transformers to \emph{learn each fact twice}. In contrast, we were unable to find evidence for this conclusing with supervised probing. Capacity scaling for two-hop QA with chain of thought suggests each fact is only learned once, consistent with the better generalization performance of this configuration. We also show that with appropriate dataset parameters, it is possible to ``trap'' very small models in a regime where they memorize answers to two-hop questions independently, even though they would perform better if they could learn to answer them with function composition. Our findings show that measurement of capacity scaling can complement existing interpretability methods.
\end{abstract}

\section{Introduction}
\label{introduction}

\subsection{Motivation}

Many existing transformer interpretability methods, like probing \citep{belinkov2022probing} and sparse autoencoders \citep{huben2023sparse} involve extracting interpretable signals from the activations. These allow us to identify features that are relevant to the computation for particular data points, but they also necessarily ignore other parts of the computation. Another ubiquitous method for understanding models is to test how they generalize to situations absent from the training data -- this is extremely useful, but so standard it would not typically be considered an ``interpretability technique''. Both methods come with limitations: hidden state probing may fail to capture important features of the computation or fail to yield useful hypotheses about how the model is computing answers, and it may be infeasible to test a model's generalization performance on all tasks of interest.

We investigate a new, complementary method for interpreting transformers: \emph{information capacity scaling}. Prior work by \citet{allen-zhuPhysicsLanguageModels2024} found that sufficiently trained transformers can memorize facts up to a limit of around 2 bits per parameter, independent of architecture or dataset size.

We hypothesize that the figure of 2 bits per parameter represents how much a transformer trained to saturation can be compressed with respect to a model of computation that matches the transformer's architecture. This compression will be less efficient than the ``ideal'' compression in tasks that benefit from recurrence, because recurrent models of computation allow for compact representation of repeated application of functions while transformers must explicitly encode each function application. We will call this imprecise notion of compression the ``native information content'' of a transformer, and we will study how this figure scales with model size for the task of two-hop question answering.

The native information content of transformers is interesting because, we suggest, it is related to how well transformers generalize. The size of an uncompressed transformer is a very loose upper bound on its information content -- an untrained transformer is the same size as a transformer trained to saturation. On the other hand, it is conceivable that a very smart compression algorithm produces a better model of the data generating process than the transformer it is compressing (for example, by exploiting recurrent patterns in the weights of the model being compressed). We speculate that the native information content represents the right level of compression for studying the generalization properties of transformers.

Information compression is closely linked to learning in general. A precise, if narrow, version of this principle is: given a sequence of data $(X_1, X_2, ...)$ sampled from a distribution $P(X)$, the better we can model $P$, the more efficiently we can encode the data. Many authors have expressed a broader view that compression and intelligvence are tightly linked \citep{deletang2023language}, often partly informed by the classical result of \citep{solomonoff1964formalI} linking representation length with predictive performance.

Practically, we make specific predictions about generalization based on specific algorithmic hypotheses and their information content scaling properties. This rests on the assumption that algorithms with the same information content scaling properties tend to have the same generalization behaviour. While the current work seems to offer modest support for this position, it is not clear how often it is applicable in general. One could also test more general hypotheses; for example, in work that tests whether transformers use correct multihop reasoning or take shortcuts (such as \citet{dziriFaithFateLimits2023}) we could measure how information capacity scaling diverges from the scaling of correct multihop reasoning, which may allow us to conclude that transformers employ less efficient algorithms than correct multihop reasoners in general, without offering specific predictions about generalization.

\subsection{Two hop question answering}

In this work we measure the information content scaling of transformers trained to answer two-hop questions. Suppose we have a dataset of two-hop questions with answers (that is, strings of the form ``Who is Bob's boss's best friend? Mary''). If we have a dataset that contains such a question for every person and pair of relationships, then a model that memorizes the answers $h(\text{Bob}, \text{boss}, \text{best friend})$ to all such questions independently will take $|N||R|^2 \log |N|$ bits (this is the ``hash table'' model of computation), where $N$ is the set of people in the dataset and $R$ the set of relationships. There is a much more efficient way to answer such questions: we learn a function $f:N\times R\to N$ that memorizes each person's relationships, taking $|N||R|\log |N|$ bits, and then we apply it twice. That is, we would answer the original question with $f(f(\text{Bob}, \text{boss}), \text{best friend})$. Because $f$ is re-used, we reduce the memorization requirement by a factor of $|R|$ -- but, because $f$ is reused, we believe this algorithm is impossible to implement without recurrence.

Because a transformer is purely feed forward, by the time an answer to $f(\text{Bob}, \text{boss})$ is obtained, it cannot ``go back'' and reapply $f$. Instead, it may need to learn two functions $f_1$ and $f_2$ with the only difference being that $f_2$ is stored in later layers than $f_1$, and compute $f_2(f_1(\text{Bob}, \text{boss}), \text{best friend})$. This is far from the only algorithm a transformer could learn that is compatible with its feed-forward architecture. The main point is that two-hop question answering requires answers (or partial answers) to factual lookups to be fed back into the same factual lookups, and this appears to necessitate that transformers replicate factual information across different layers.

\subsubsection{Generalization properties of different two-hop algorithms}

We represent two-hop questions using the schema $(e_1, r, e_2, a, s)$, where $e_1$ is the first entity (in our case, entities are people), $r$ is the first relation, $e_2$ is the intermediate entity, $a$ is the attribute in question (which may be another relation, or some other property of $e_2$), and $s$ is the solution. For example, in the question ``Who is Bob's mother's boss?'', we have $e_1=\text{Bob}$, $r=\text{mother}$, $e_2=\text{Bob's mother}$, $a=\text{boss}$, and $s=\text{Mary}$. Note that specifying $(e_1, r, a)$ is sufficient to fully specify a two-hop question.

If we suppose that the training data only determines the content of the ``fact lookup functions'' ($h, f, f_1, f_2$), and any ``wrapper code'' required to use them is learned perfectly in every case, we can deduce a different generalization signature for each of the three schemes presented. The hash table model $h$ will only correctly answer questions that are present in the training data, while ``two function composition'' $f_1, f_2$ requires $(e_1, r)$ present \emph{as a first hop} in training so it is learned by $f_1$, and $(e_2, a)$ present \emph{as a second hop} in training so it is learned by $f_2$. Finally, ``recurrent composition'' with $f$ requires only that both facts $(e_1, r)$ and $(e_2, a)$ are present in training - once they are learned by $f$, they can be applied in either position. This is summarized in Table \ref{tab:generalization_props}.

\begin{table*}[h]
\centering
\begin{tabular}{lccccc}
\toprule
\multirow{2}{*}{Algorithm} & \multicolumn{3}{c}{Present in Training} & \multirow{2}{*}{Correctly answered} \\
\cmidrule(lr){2-4}
& $(e_1, r)$ and $(e_2, a)$ & First Hop $(e_1, r)$ and Second Hop $(e_2, a)$ & $(e_1, r, a)$ & $(e_1, r, a)$ \\
\midrule
$h(e_1, r, a)$ & Yes & Yes & Yes & Yes \\
$h(e_1, r, a)$ & Yes & Yes & No & No \\
$h(e_1, r, a)$ & Yes & No & No & No \\
$f_2(f_1(e_1, r), a)$ & Yes & Yes & Yes & Yes \\
$f_2(f_1(e_1, r), a)$ & Yes & Yes & No & Yes \\
$f_2(f_1(e_1, r), a)$ & Yes & No & No & No \\
$f(f(e_1, r), a)$ & Yes & Yes & Yes & Yes \\
$f(f(e_1, r), a)$ & Yes & Yes & No & Yes \\
$f(f(e_1, r), a)$ & Yes & No & No & Yes \\
\bottomrule
\end{tabular}
\caption{Predicted generalization properties of different algorithms for two-hop question answering.\label{tab:generalization_props}}
\end{table*}

\subsubsection{Existing work on multi-hop reasoning in Transformers}

\citet{wangGrokkedTransformersAre2024} found evidence for the ``two function composition'' algorithm for answering latent two-hop questions. Specifically, they found:

\begin{itemize}
    \item Supervised probes in the middle layers can extract the answers to the first hop of two-hop questions in transformers trained to answer two-hop questions
    \item Transformers trained to answer two-hop questions fail to generalize to unseen questions $(e_1, r_1, r_2)$ if $(e_1, r_1)$ and $(e_2, r_2)$ are only seen in the context of one-hop questions in the training data
\end{itemize}

In contrast, \citet{yuLLMsReallyThink2025a} using similar probing methods on a dataset of multi-step arithmetic problems found that intermediate steps were often non-recoverable with probing prompted models, even when these models could answer the questions with nontrivial accuracy, though they found fine-tuned models allowed for intermediate step recovery with probing.

Our own probing results are weaker than these: in fact, we failed to recover intermediate answers using probing (see Section \ref{sec:probing}) even when these models had generalization results and capacities that matched the two function composition hypothesis. That is, in our experiments, inferences based on information capacity matched generalization results better than inferences based on probing results.

Transformers are known to be weak at compositional tasks in general. \citet{dziriFaithFateLimits2023} found that transformers often failed to learn to generalize to compositional tasks whose graphical representation was not seen in training data. Theoretical analyses have also placed limits on the capabilities of transformers without chain of thought \citep{merrillSaturatedTransformersAre2022, liuTransformersLearnShortcuts2023}. These limitations are not present for transformers that may use chain of thought reasoning \citep{perezAttentionTuringComplete2021}.

\subsection{Summary}

\begin{itemize}
    \item We trained transformers of a variety of sizes and depths on a collection of synthetic two-hop QA datasets of varying sizes
    \item For one-hop QA, we approximately reproduced previous results of a ``knowledge capacity'' of about 2 bits per parameter, though our measured capacities were somewhat lower
    \item For two-hop QA, we found that the ``two function composition'' algorithm best fit the hypothesis of a 2 bits per parameter knowledge capacity
    \item We did not find evidence for two function composition using supervised probing
    \item We conclude that knowledge capacity measurement is an alternative method to investigate questions about which algorithms a transformer implements which can be informative where probing fails, but it comes with significant challenges
    \item We also have a number of minor results:
    \begin{itemize}
        \item We were able to ``trap'' some small transformers in a regime where they memorized the answers to every two-hop question independently instead of learning to compose functions
        \item We conduct a more comprehensive assessment of generalization in two-hop QA than previous work, finding that in every case the result appears consistent with ``two function composition''
    \end{itemize}
\end{itemize}

\section{Method}

\subsection{Data}

We generated datasets of fictional profiles of ``people'' from $|N|=1000$ to $|N|=250 000$ profiles. We call the set of people $N$. Each person had a randomly selected first, middle and last name, with $|N_0|=4\times 10^{11}$ possible combinations. Name combinations were selected without replacement.

Each person had 17 types of ``relations'' to other people. To make the problem easy to analyse, the relations had no structure - while some relations were called ``parents'' and others ``children'', these relations were independently uniformly randomly selected, so the ``child'' relation is \emph{not} the inverse of the ``parent'' relation. We call the set of relations $R$. Each person also had 4 ``properties''. These could be queried at the last hop, but could not constitute a first hop (as the possible values were not people). Call the set of relations and properties ``attributes'', $A$. Each attribute $a_j\in A$ has a set of possible values $V_j$.

We had 21 different attributes: 17 relations and 4 properties (birth city, birth date, employer and university).

The profiles were then used to generate questions from templates: one-hop questions took the form ``What was \{person\}'s \{relation/attribute\}? \{answer\}'', while two-hop questions took the form ``What was \{person\}'s \{relation\}'s \{relation/attribute\}? \{answer\}''. An earlier version of this work employed diverse paraphrases, but this was abandoned to make the experiment simpler. Thus while the training data is English text, it does not feature any of the variety or structure typically associated with English text.

In every case, all facts were included in the dataset as one-hop questions. We created 7 different held out sets of two-hop questions to test generalization under different conditions. To create hold out sets, we selected particular components of two-hop questions and removed every two-hop question from the training data that featured this component. The held out components were:
\begin{itemize}
    \item First Entities ($e_1$)
    \item First Relations ($r$)
    \item Second Entities ($e_2$)
    \item Attributes ($a$)
    \item Entity 1-relation pairs $(e_1, r)$
    \item Entity 2-attribute pairs $(e_2, a)$
    \item Complete questions $(e_1, r, a)$
\end{itemize}
For example, holding out the relation ``mother'' means that we remove all two-hop questions containing the mother relation, but one-hop questions containing ``mother'' are still present in the training data.

\subsection{Training}

We trained transformers based on the Llama architecture \citep{dubey2024llama} from 250K to 15M parameters. We employed a custom tokenizer trained on our data with a vocabulary of 3K in order to facilitate very small parameter counts, since embedding parameters would take up a large fraction of the total parameters otherwise.

We used the $\mu$P parametrization \citep{yangTensorProgramsTuning2022} for our transformers to enable proper transfer of hyperparameters across different model widths. We used the schedule free AdamW optimizer \citep{defazio2024road}, which allowed us to train for as long as was necessary to achieve our convergence criterion without specifying a learning rate schedule in advance. All tokens but the answer tokens were masked.

A batch size of 32 sequences was used to train all models, and training text was split into 500 token chunks, so that each batch contained 16K tokens (including special tokens). We trained until the loss decreased by less than $10^{-8}$ per step, this typically took about 10M steps (or 160B tokens) and did not depend strongly on the size of the dataset.

Unless we were training on one-hop questions only, models were trained on a single one-hop question for every 10 two-hop questions.

\subsection{Information Content Measurement}

By \emph{dataset entropy}, we mean the entropy of a random variable $X:=(X_1, X_2, X_3,...)$ such that our entire dataset is considered one sample of it.

We  define such a random variable as 
\begin{align*}
    X=(X_{\text{names}}, X_{\text{birth dates}}, X_{e_1 a_1}, ... ,X_{e_1 a_{|A|}}, ... ,X_{e_{|N|} a_{|A|}})
\end{align*}
Where $X_{\text{names}}$ is a random variable describing the uniform selection of $N$ names from $N_0$ possibilities, and $X_{\text{birth dates}}$ a random variable describing the selection of $|N|$ birth dates from $N_b$ possibilities. Conditioned on these two variables, the attribute values $X_{e_i a_j}$ are uniform IID:
\begin{align*}
    P(X_{e_i a_j}|X_{e_{N\setminus i} a_{|A|\setminus j}} X_{\text{names}}, X_{\text{birth dates}}) = \frac{1}{|V_j|}
\end{align*}
Here we assume that the receiver already knows which tokens may be used for each answer type, though they don't know, for example, the actual set of names in the dataset. We neglected terms related to the selection of viable tokens from all possible tokens because this number does not scale with the total number of profiles. There are other terms that we can't easily quantify that also don't scale with the number of profiles (e.g. learning to attend to the relevant tokens in the question), and we therefore opted to focus on the scaling with the dataset size.

However the entropy of selecting $N$ names from $N_0$ possibilities does scale with $N$. We considered this to contribute $N\log_2 N_0$ bits to the dataset entropy.\footnote{Strictly speaking, there are $\binom{N}{N_0}$ ways to select an unordered set of $N$ names without replacement from $N_0$ possibilities, corresponding to $\log_2 \binom{N}{N_0}$ bits of information. But in our case $|N| \ll N_0$, so this is a close approximation.} %This is the number of bits needed to specify $N$ selections from $N_0$ possibilities with replacement (we didn't consider no replacement because $N_0 \gg N$) if order matters. Order does not matter, but (TODO: this seemed to fit some early results better and doing a systematic study never rose up high enough on the list of priorities).

For one-hop QA, the entropy of an individual answer for an attribute $a$ is $\log_2 |V_a|$, where $V_a$ is the set of possible answers for attribute $a$, and so the dataset entropy is:
\begin{align}
    E_1 := |N|\log_2 N_0 + |N|\sum_{a\in A\cup R} \log_2 |V_a|
\end{align}
Here, $A$ is the set of available attributes and $R$ is the set of available relations.

\begin{figure}[ht]
    \centering
    \includegraphics[width=1\linewidth]{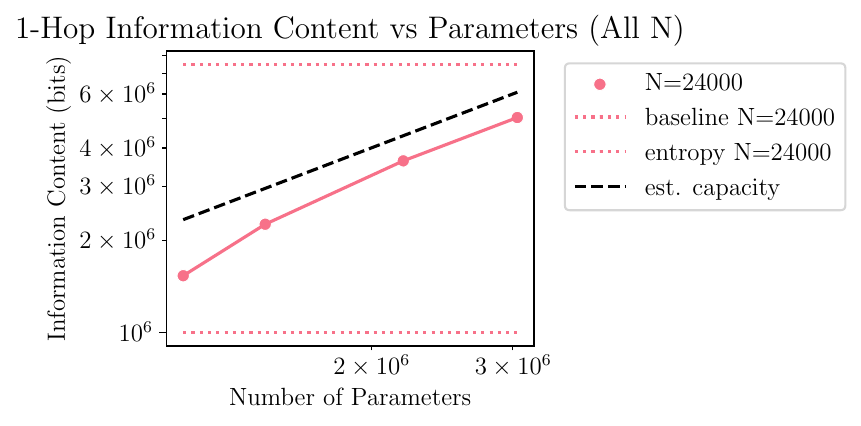}
    \caption{Observed information content scaling on one-hop questions for 4 layer transformers. A loss of zero implies that the information content is equal to the dataset entropy, and a loss equal to predicting the uniform distribution over all answers in the dataset yields the information content level represented by ``baseline'' (this is nontrivial due to needing to learn the set of names in the dataset out of all possible names). The dashed black line represents a content of 2 bits per parameter.}
    \label{fig:one-hop-scaling}
\end{figure}

For two-hop QA, we considered a number of different schemes. In the ``recurrent'' computational model, the dataset entropy is the same as the one-hop entropy; one can answer all two-hop questions using the recurrent model iff one knows all of the one-hop relations.
\begin{align}
    E_2^{\text{recurrent}}:=E_1
\end{align}
Next, we consider the ``two-function'' computational model. Under this model, ``first hop'' relations and ``second hop'' relations are independent facts and therefore have to be sent twice:
\begin{align}
    E_2^{\text{2f}}:= |N|\log_2 N_0 + 2N\sum_{a\in A} \log_2 |V_a|
\end{align}
Finally, in the ``independent'' model, every answer to every two-hop question is treated as independent. The entropy is equivalent to the entropy of a one-hop dataset with $|R|N$ entities:
\begin{align}
    E_2^\text{independent}=|N|\log_2 N_0 + |R||N|\sum_{a\in A} \log_2 |V_a|
\end{align}

To measure the information content of a trained model, for a one-hop QA model we take the difference between the dataset entropy and the sum of cross-entropy losses over all answers and all tokens in each answer (see \citet{allen-zhuPhysicsLanguageModels2024}):
\begin{align}
    C_1 \geq E_1 + |N||A|\mathbb{E}\left[\log_2 p_\text{one-hop}\right]\label{eq:1hop_content}
\end{align}
Here $p_\text{one-hop}$ is the probability of a correct one-hop answer (across all answer tokens), and the expectation is taken over the entire dataset.
We employ a similar rule for independent two-hop QA:
\begin{align}
    C_2^\text{independent} \geq E_2^{\text{independent}} - |N||R||A|\mathbb{E}\left[\log_2 p_\text{two-hop}\right]
\end{align}
For recurrent composition and two-function composition, we calculate an ``effective loss'' for the one-hop functions that are composed to yield the two-hop answer, details are in Appendices \ref{ssec:two_hop_recurrent} and \ref{ssec:two_hop_two_fn}. We then employ Equation \ref{eq:1hop_content} to compute the content:
\begin{align}
    C_2^{\text{recurrent}} \geq &E_2^\text{recurrent} - |N||A|\mathbb{E} \left[\log_2 p_\text{eff}^\text{recurrent}\right]\\
    C_2^{\text{2f}} \geq &E_2^\text{2f} - |N||A|(\mathbb{E} \left[\log_2 p_\text{eff}^\text{hop 1}\right] + \mathbb{E} \left[\log_2 p_\text{eff}^\text{hop 2}\right])
\end{align}

\section{Results}

\subsection{Information Content}

We did not perfectly reproduce the existing result of a 2 bit per parameter information capacity for one-hop questions and answers. Instead, \textbf{our one-hop training runs yielded an information capacity of around 1.6 bits per parameter} (Figure \ref{fig:one-hop-scaling}). We did successfully replicate the figure in earlier training runs (Figure \ref{fig:1-hop-old}), but these runs had different hyperparameters to the two-hop training runs that appear in this paper -- they did not employ $\mu$P parametrization and used 4 relations instead of 17 relations.

\begin{figure}[t]
    \centering
    \includegraphics[width=1\linewidth]{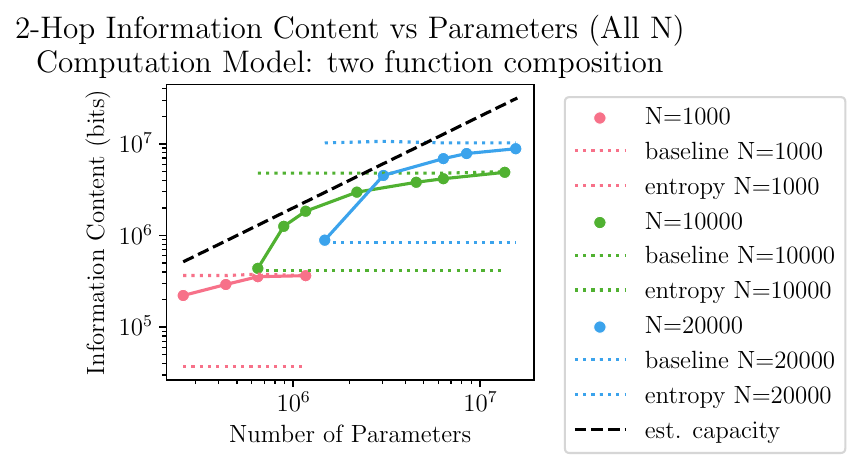}w
    \caption{Observed information content scaling on two-hop questions without chain-of-thought generation for 4 layer transformers, assuming the two-function composition computational model. The dataset entropy and baseline curves are not quite constant due to randomly held out attributes differing between datasets.}
    \label{fig:2hop-scaling}
\end{figure}

We observed that transformers exhibited ``stacked curve'' information content scaling for different dataset sizes (Figure \ref{fig:2hop-scaling}). These curves generally began far below the hypothesized 2 bit per parameter capacity, and such transformers had not learned much more than uniform guessing of the answers. Subsequently, the content curve almost kisses the 2 bit per parameter line, before leveling off as it approaches the dataset entropy.

Under the 2 bits per parameter capacity hypothesis, these results fit the two function composition computational model far better than they fit either of the alternative models. This certainly isn't conclusive evidence for the hypotheses: we were only able to assess a limited range of dataset and parameter sizes, we are testing two hypotheses at once (2 bits per parameter capacity as well as the computational model) and the scaling behaviour is more complex than identified in \citet{allen-zhuPhysicsLanguageModels2024} - here, the scaling curves seem to bend towards both the baseline and dataset entropy lines. It is, however, stronger support for the main hypothesis than any of the identified alternatives.

The curved dependence of information content on parameter count may be due to a few factors: first, when the models' performance on either individual hop (or both together) is too weak, it may be difficult to learn the composition algorithm. Second, models may choose to allocate some fraction of their information budget to memorization, and this fraction may increase as they approach perfect train set performance. Because they are still near their capacity limits, allocating budget to inefficient memorization may prevent them from achieving zero loss on the data. This may be weakly supported by the observation that the ``generalization gap'' (that is, the difference between train loss and loss on held out questions) grows with larger parameter counts (Figure~\ref{fig:generalization-gap}), though this trend may reverse with sufficiently high parameter counts, see the final data point for $N=1000$. Finally, we use a very crude approximation for the measuring the two-hop capacity when the variance in per-question loss is large, and it may be the case that this is responsible for the shape of the curve.

\citet{allen-zhuPhysicsLanguageModels2024} noted that information capacity did not seem to depend on the model architecture. We also trained deeper models -- 12 layers rather than 4 -- and observed a similar but less consistent trend (Figure \ref{fig:2-hop-deep}). This may be due to the fact that we tuned optimizer hyperparameters for the shallower model and did not adjust them for the deeper model.

\begin{figure}
    \centering
    \includegraphics[width=1\linewidth]{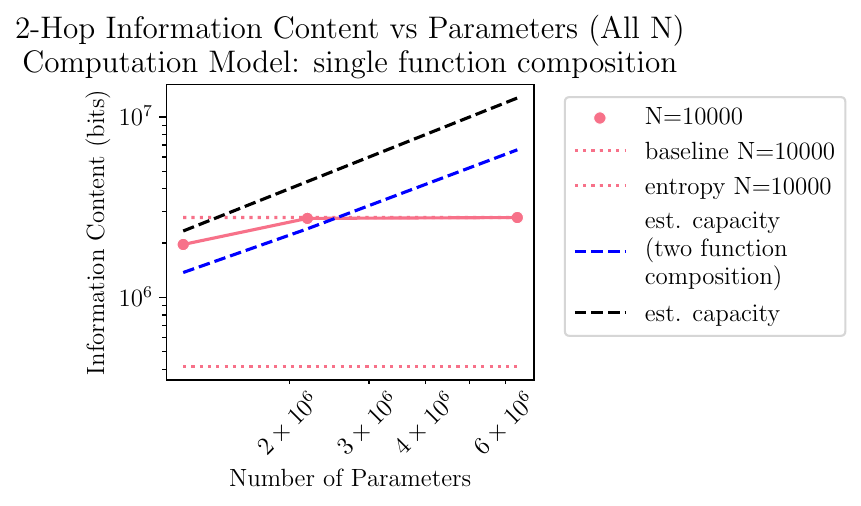}
    \caption{Observed information content scaling on two-hop questions with chain-of-thought generation for 4 layer transformers. The information content exceeds the capacity estimate for two function composition and approaches the capacity for recurrent composition, expected.}
    \label{fig:2hop-scaling-cot}
\end{figure}

\subsection{Two Hop Generalization}

\begin{figure}
    \centering
    \includegraphics[width=1\linewidth]{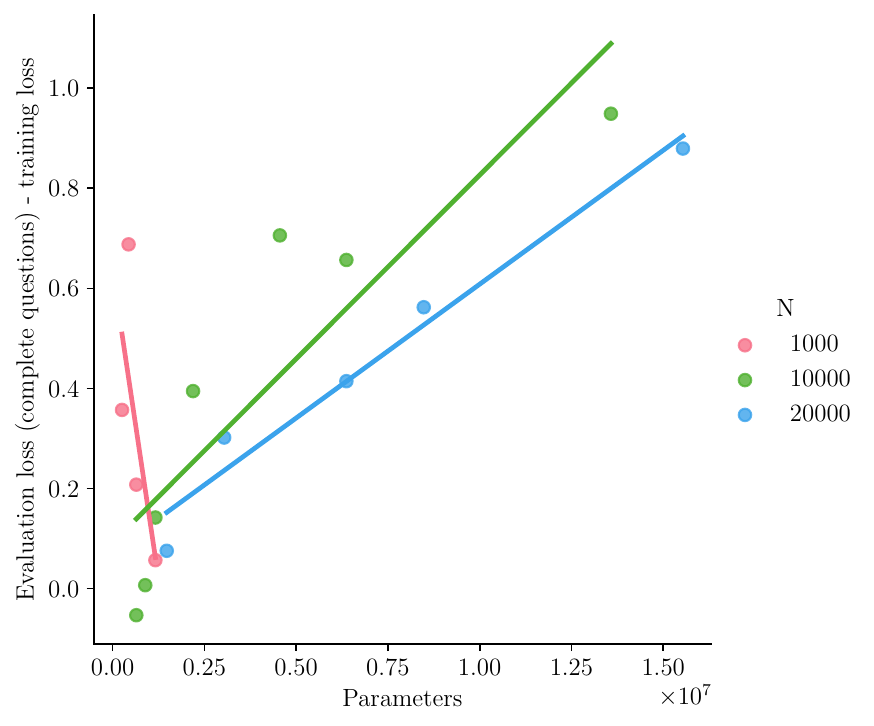}
    \caption{The generalization gap for two-hop question answering. In most cases the gap increased as the parameter count increased. Models trained on the 1000 profile dataset were an exception, with the largest model on this dataset achieving nearly 0 loss on both the train and evaluation sets.}
    \label{fig:generalization-gap}
\end{figure}

We find the generalization performance of transformers with no chain-of-thought on two-hop problems precisely matches the prediction of Table \ref{tab:generalization_props}. In short: without chain of thought, there was no generalization to any case where any component of the first or second hops was systematically excluded from the training data, but there was generalization to the case where complete questions were excluded from the training data (Figures \ref{fig:random_guessing_generalization} and \ref{fig:generalization-gap} respectively). This matches prior work by \citet{wangGrokkedTransformersAre2024} and extends it by systematically excluding different parts of each hop of two-hop questions, whereas prior work systematically excluded facts from appearing anywhere in any two-hop question.

Recall that training always presented every fact in the context of one-hop questions. Thus our work indicates that facts learned in the context of one-hop questions do not get learned as components of ``$f_1$'' or ``$f_2$'' unless they actually appear in the right position in a two-hop question.

The situation was somewhat more complicated for generalization with chain-of-thought. We never saw generalization to held out first entities, and always to held out entity 2-attribute pairs, but other held out elements showed generalization in some training runs but not others without a clear pattern (Figure \ref{fig:random_guessing_generalization_cot}. For generalization in chain-of-thought generations the model had to produce tokens (or sets of tokens) that were never seen in the same position during training; we speculate that the inconsistent generalization was the result of competing heuristics (i.e. low marginal probabilities of tokens appearing in some position vs applying the correct rules to answer a two-hop question). While we did not train many models of this type, it's possible that generalization on particular held out sets for this setup is seed dependent \citep{zhangInitializationCriticalWhether2025}. This does indicate that scaling -- which was consistent with recurrent composition -- is not a perfect proxy for generalization.

\subsection{Two Hop independent memorization}

We have speculated that models learn both independent memorization and the generalizing two function composition algorithms to carry out two-hop reasoning. We experimented with an alternative training regime, where models were incentivised to memorize two-hop answers independently at the expense of learning one-hop answers. We found that these models never learned to generalize, and exhibited significantly lower information content than generalizing models.

\begin{figure}
    \centering
    \includegraphics[width=1\linewidth]{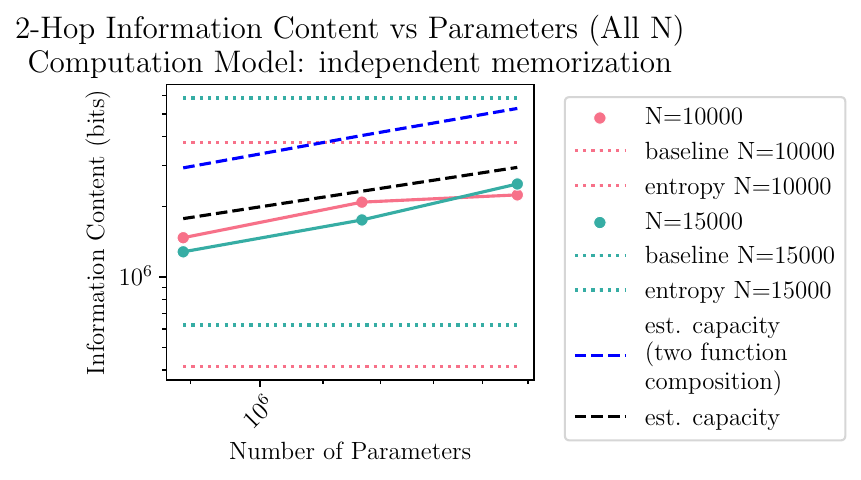}
    \caption{Anomalously low information content for models trained with only 4 relations. The estimated information content approximates the capacity curve if we assume independent memorization of all two hop questions.}
    \label{fig:two-hop-scalining-nogeneralization}
\end{figure}

Specifically, we used a dataset where each person had 4 relations and 4 properties, rather than 17 relations, and we maintained the ratio of 1 one-hop question for every 10 two-hop questions. In this case, reducing the loss for an individual two-hop question by a fixed amount has $\frac{10}{4}=2.5$ times the impact on the sum of loss for all questions as reducing the loss for an individual one-hop question. We observe that these models do not learn one-hop question answering much beyond the uniform distribution on possible answers, their capacity scaling approximately matches the expected scaling for independent memorization (Figure \ref{fig:two-hop-scalining-nogeneralization}) and they do not generalize to any held out questions at all.

We hypothesize that the mechanism at work in this case is that the model first learns to memorize individual two-hop answers, because this has a larger impact on the average loss than memorizing one-hop answers, and transformers tend to learn simple rules first (see, for example \citet{belroseneural}). However, due to the limited capacity of our small models, they continue memorizing answers until they do not have capacity to memorize further. This leads the models being trapped in a local minimum: improving performance on one-hop questions (which is necessary to learn function composition) requires making performance worse on the memorized two-hop questions due to capacity limitations. Thus they are ``trapped'' having learned the nongeneralizing function, even though in principle they have the capacity to perform substantially better if they learned to generalize.

\subsection{Probing}\label{sec:probing}

\citet{wangGrokkedTransformersAre2024} used a \emph{logit lens} to show that in their generalising two-hop reasoning circuits, the intermediate entities could be recovered by probing the appropriate token positions. Here we employ supervised linear probes to try to recover intermediate entities at a range of token positions. We expect supervised linear probes to be more effective at extracting linearly retrievable information than the logit lens, which simply applies the model's unembedding weights to the hidden layer activations without any training on the hidden layer specifically.

Somewhat surprisingly, and in contrast to our capacity investigations, we did not find a signal that strongly differentiated between generalizing and nongeneralizing models. See Tables \ref{tab:loss_metrics} and \ref{tab:loss_metrics_alt}. In particular, these results show that the intermediate entity is not more retrievable in models where we infer two-function composition than in models where we infer memorization, and in fact the second entity is barely more reterievable than an arbitrary relation of the first entity.

We have compelling evidence of systematic differences between memorizing models and models that implement two function composition -- they generalize differently, and exhibit different capacities for learning two hop QA -- but these differences were not apparent to supervised probes.

\subsection{Discussion and Conclusion}

We set out to explore the viability of using information content measurement as a complementary method of transformer interpretability. We adopted a hypothesis driven approach  -- we guessed that transformers might implement a certain kind of function, deduced information content scaling principles for this kind of function, and tested whether the capacity for transformers to learn the dataset corresponded to these scaling principles.

There are significant challenges to be overcome in order to apply this to transformer interpretability in a scalable fashion. Even in relatively simple cases, it is nontrivial to go from a hypothesis to a capacity scaling curve, and the kinds of transformers we are most interested in are trained on complex and poorly understood data generating processes. We find that we do not reproduce the precise 2 bits per parameter empirical scaling curve on simple factual lookup data, and the fact that dataset choices may introduce uncertainty in this curve makes hypothesis testing more difficult.

Despite this, we do find some support for our main hypothesis that transformers learn two-hop reasoning by memorizing all relevant facts twice by comparing empirical information contents to the 2 bits per parameter capacity figure. We also find, as expected, that models using chain of thought are much more efficient at learning two-hop reasoning. Finally, we saw that models with unusually low information content did not generalize, supporting the hypothesis that they learned an inefficient and non-generalizing representation of the data. 

A possible real-world example of information content efficiency is the strong performance in mathematical benchmarks for very small distilled ``reasoning models'' found by \citet{deepseek-aiDeepSeekR1IncentivizingReasoning2025}. Reasoning models are trained to rely much more on chain-of-thought than traditional language models. As our results suggest, transformers may have a much higher capacity to learn sequential reasoning with chain of thought than in a single forward pass, which seems to require redundant memorization of known facts. It is not clear whether non-reasoning models would actually be running into hard capacity limits here, and it is likely to be difficult to test.

While it is not straightforward to apply this method to interpret models, it is plausible that it can be made to work if the application is compelling enough. The question is whether there are sufficiently compelling applications that cannot be addressed by other methods. At this stage, we are not sure if there are many such applications.

\section*{Contributions and Acknowledgements}

David Johnston came up with the main ideas behind this project, performed experiments and analysis. Sergio performed additional experiments and contributed analyses. Nora Belrose gave advice and guidance throughout the project. David and Nora are funded by a \href{https://www.openphilanthropy.org/grants/eleuther-ai-interpretability-research/}{grant} from Open Philanthropy. Sergio contributed as a volunteer. We thank Coreweave for computing resources.

\bibliography{citations}
\bibliographystyle{icml2025}

\newpage
\appendix
\onecolumn

\section{Information content estimation}

\subsection{Two hop recurrent function composition}\label{ssec:two_hop_recurrent}

To simplify the following discussion, we consider only the relations that entities possess and not the additional properties.

Consider the case where we have a collection of two-hop questions and answers. We know that these are generated deterministically from a function $f$ mapping one-hop questions to answers. There may be several such functions that lead to the same two-hop questions and answers: if $\forall j: f(e_i, r_j)=f(e_k, r_j)$ then defining $g$ such that $g(e_l, r_m)=f(e_l, r_m)$ unless $f(e_l, r_m)=e_i$, in which case $g(e_l,r_m)=e_k$ will lead to the same two-hop questions and answers (i.e. if two intermediate entities have identical signatures, then they can be swapped). This is extremely unlikely in our generation scheme - we have $N$ entities, with the probability of a perfect match between any two less than $\frac{1}{N^{17}}$, which is very small compared to $\binom{N}{2}$.

There are unlikely to be other symmetries of $f$ with respect to two-hop question answering. If we want $g(e_i,r_j)=e_k\neq f(e_i, r_j)=e_p$ such that $f(e_p,r_l)\neq f(e_k, r_l)$ for some $l$, then we require $\forall m: g(e_k, r_m) = f(e_k, r_m)$ (to maintain the same two-hop answers), and also $\forall n: g(e_q, r_n) = f(e_p, r_n)$ for some $q$ (because $e_p$ is an intermediate entity in some two-hop questions). If $e_q$ does not have the same relations as $e_p$ (which would be very unlikely in our setting), we need an additional entity to play the role of $e_p$ and so on. It is very likely that this exploding set of constraints eventually leads to a collision with the requirement to answer two-hop questions correctly: if, for example, $g(e_k, r_m) = f(e_k, r_m) = e_q$ then we require both that $\forall n : g(e_q, r_n) = f(e_p, r_n)$ but also $\forall n : g(e_q, r_n) = f(e_q, r_n)$ to keep the answers to the two-hop questions the same -- which, as we've already argued, is very unlikely.

Thus having $f$ is sufficient to construct the set of all two-hop questions and it is also overwhelmingly likely to be necessary. For this reason, we treat the entropy of $f$ -- understood as a random variable whose distribution is given by our data generating process -- to be equal to the entropy of the two-hop question and answer dataset.

Given this assumption, we can lower bound the information content of a recurrent function composer based on the loss of its answers to two-hop questions. The model reconstructs a one-hop probability $P_1$ such that

\begin{align*}
    P_2(e_{112}^*|e_1, r_1, r_2) = P_1(e_{11}^*|e_1, r_1)P_1(e_{112}^*|e_{11}^*, r_2) + \frac{1-P_1(e_{11}^*|e_1,r_2)}{|N|}
\end{align*}

Where $e_{11}^* = f(e_1,r_1)$ and $e_{112}^* = f(e_2, r_2)$. Let $q_{112}:= P_2(e_{112}^*|e_1, r_1, r_2)$, $p_{11}:=P_1(e_{11}^*|e_1, r_1)$ and $p_{112}=P_1(e_{112}^*|e_{11^*},r_2)$. We assume the $p_{ij}$s are mutually independent.\footnote{The $p_{ij}$s may in fact be negatively correlated because the model is spreading a total budget across all $i$, $j$ pairs, but $|N||R|$ is large and so this correlation will be small.}

\begin{align*}
    \mathbb{E}_{i,j,k\sim U(N\times R^2)}[q_{ijk}] &= \mathbb{E}_{i,j\sim U(N\times R)}[p_{ij}]^2 + \mathbb{E}_{i,j\sim U(N\times R)}\left[\frac{1-p_{ij}}{|N|}\right]\\
    \mathbb{E}[p_{ij}] &= \frac{-1 + \sqrt{1- 4|N|(1 - |N|\mathbb{E}[q_{ijk}])}}{2|N|}
\end{align*}
where we have taken the result between 0 and 1.

We are interested in the negative sum of log probabilities.
\begin{align*}
    \log \mathbb{E}[p_{ij}] &= \log \left(\frac{-1 + \sqrt{1- 4|N|(1 - |N|\mathbb{E}[q_{ijk}])}}{2|N|}\right)
\end{align*}

We make the approximation that
\begin{align*}
    \mathbb{E}[\log p_{ij}] &\approx \log \left(\frac{-1 + \sqrt{1- 4|N|(1 - |N|e^{\mathbb{E}[\log q_{ijk}]})}}{2|N|}\right)
\end{align*}

Empirically, when $\mathbb{E}[p_{ij}]\approx \frac{1}{N}$, $\text{Var}[\log q_{ijk}]\approx 0$ and the approximation is almost exact. On the other hand, if $\mathbb{E}[p_{ij}]]\gg \frac{1}{N}$, then

\begin{align*}
    \mathbb{E}[q_{ijk}] &\approx \mathbb{E}[p_{ij}]^2\\
    \mathbb{E}[\log p_{ij}] &\approx \frac{\mathbb{E}[\log q_{ijk}]}{2}\\
    &\approx \log \left(\frac{-1 + \sqrt{1- 4|N|(1 - |N|e^{\mathbb{E}\log q_{ijk}})}}{2|N|}\right)
\end{align*}

thus it is a close approximation at both extremes of the values of $p_{ij}$.

\subsection{Two hop two function composition}\label{ssec:two_hop_two_fn}

Recall that two function composition answers two-hop questions using a pair of functions $f_1$ and $f_2$ such that $s_{ijk}=f_2(f_1(e_i, r_j), r_k)$. This has a symmetry that is absent from recurrent function composition: if we permute all of the results of $f_1$ and apply the same permutation to the input entities to $f_2$, we preserve the same two-hop questions and answers (in the recurrent case, we cannot make these permutations independently). This means that the entropy of two function composition is approximately $|N|\log|N|$ less than double the entropy of one function composition though, because we train on one-hop question and answers as well, models probably cannot actually exploit this symmetry if they use the memorized facts $f_1$ and $f_2$ to answer one-hop questions.

Finding a lower bound on two function composition information content is similar to the recurrent function composition case, except we now have two independent ``one-hop probabilities'':

\begin{align*}
        \mathbb{E}[q_{ijk}] &= \mathbb{E}\left[p_{ij}^\text{hop 1}p_{(ij)k}^\text{hop 2} + \frac{1-p_{ij}}{|N|}\right]
\end{align*}

we assume that the model has a fixed budget that is split between first and second hops, and one unit of budget yields a fixed change in the expected log probability, in keeping with our overarching assumption about the connection between model budgets and loss. Thus we reparametrize: $u^\text{hop 1} := \mathbb{E}[p^\text{hop 1}_{ij}]$, $u^\text{hop 2} := \mathbb{E}[p^\text{hop 2}_{ij}]$, $\epsilon:= \sqrt{\frac{u^{\text{hop 1}}}{u^{\text{hop 2}}}}$, $u:= \frac{u^\text{hop 1}}{\epsilon}$.

\begin{align*}
        \mathbb{E}[q_{ijk}] &= \mathbb{E}\left[u^2 + \frac{1-u\epsilon}{|N|}\right]
\end{align*}

Choosing $\epsilon$ to maximise the right hand side gives us the minimal $u$ that satisfies this equation, and the right hand side is clearly maximised when $\epsilon$ is minimized. 

The minimal feasible value of $\epsilon$ is either the value that sets $u^\text{hop 1}$ to $\frac{1}{|N|}$ (because the model ``already knows'' that the correct answer has at least this probability) or that sets $u^\text{hop 2}$ to 1. The latter is preferred if $u^2 >\frac{1}{|N|}$, the former otherwise.

Thus, following the derivation above, and noting that we want the summed losses for both hops, we have

\begin{align*}
    -\mathbb{E} [\log p^\text{hop 1}_{ij} + \log p^\text{hop 2}_{ij}] &\approx \begin{cases}
            -2\sqrt {\frac{e^{-\mathbb{E}[\text{loss}]}(1+\text{Var}[\text{loss}]/2)-\frac{1}{|N|}}{1-\frac{1}{|N|}}} & (u^2>\frac{1}{|N|})\\
            -2\sqrt{e^{-\mathbb{E}[\text{loss}]}(1+\text{Var}[\text{loss}]/2)-\frac{1}{|N|}\left(1-\frac{1}{|N|}\right)} & (u^2\leq \frac{1}{|N|})
    \end{cases}
\end{align*}

this is our effective loss for information content calculations.

\begin{minipage}{\textwidth}
\section{Generalization to held out hop elements}
\begin{figure}[H]
    \centering
    \includegraphics[width=0.75\linewidth]{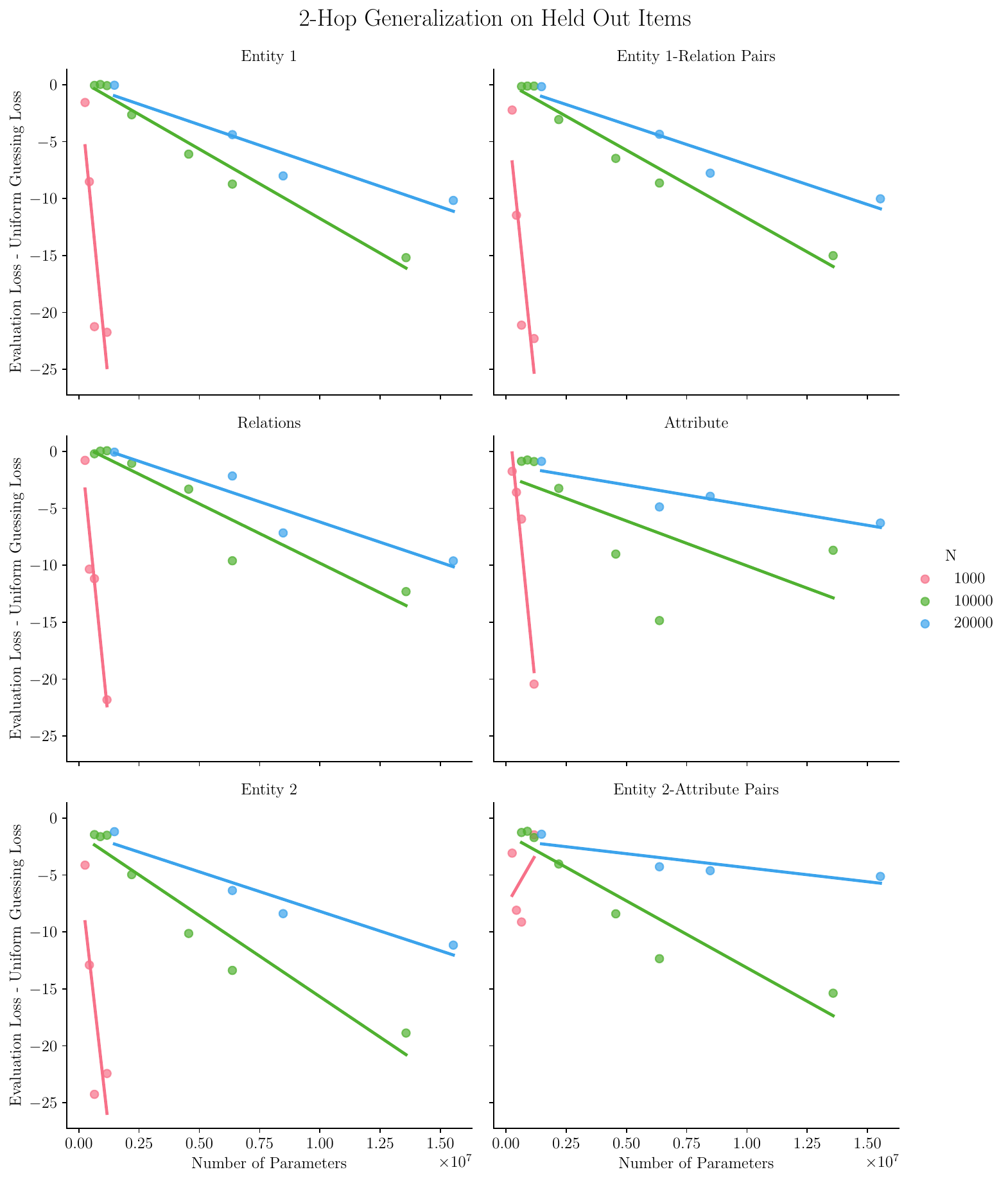}
    \caption{Comparison of uniform distribution loss with generalization loss for models answering two hop questions questions \emph{without} chain of thought with held out components. Values equal to or less than 0 indicate no generalization; in fact, every model in this plot does not generalize on any held out component.}
    \label{fig:random_guessing_generalization}
\end{figure}
\end{minipage}

\begin{figure}[H]
    \centering
    \includegraphics[width=0.75\linewidth]{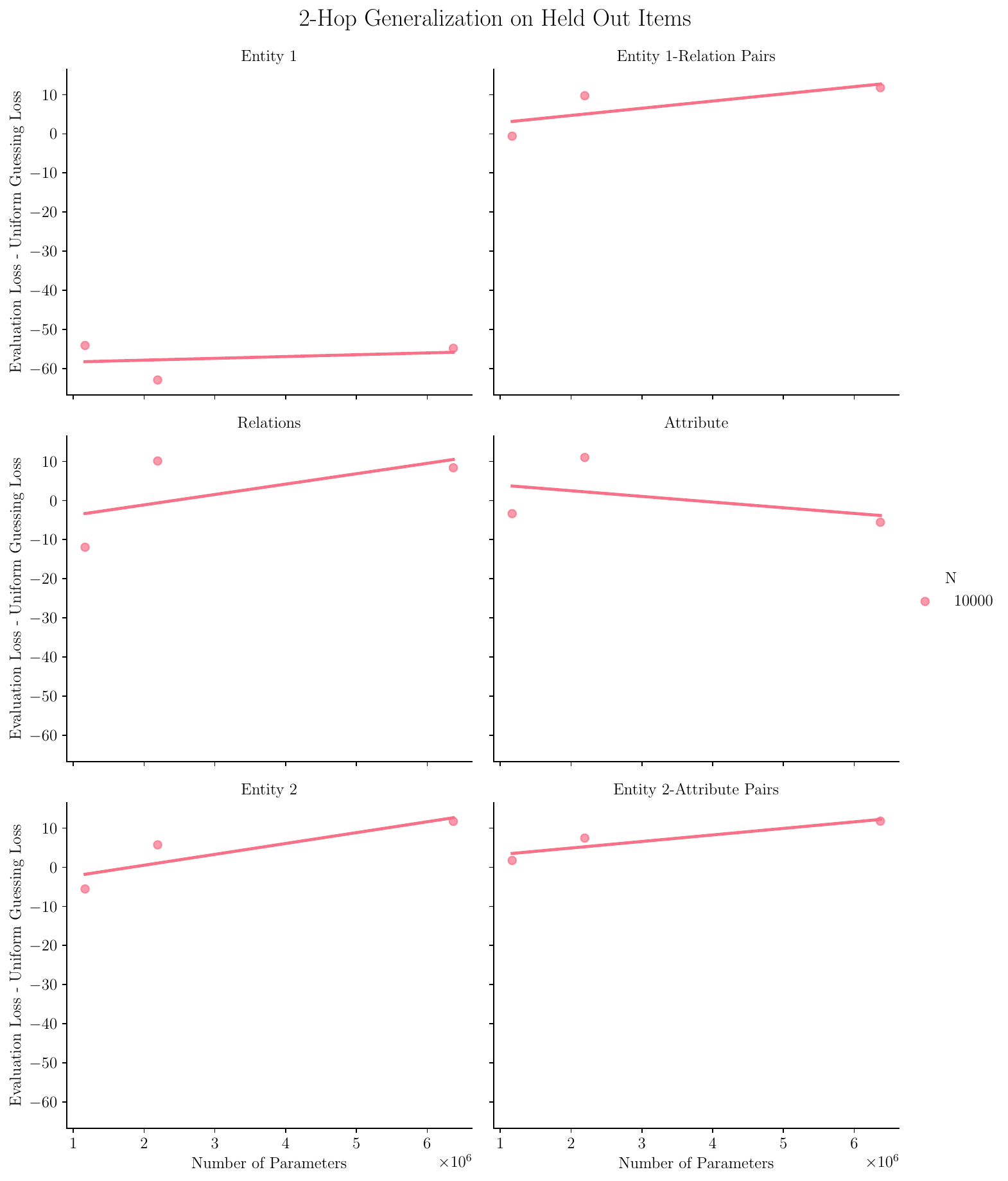}
    \caption{Comparison of uniform distribution loss with generalization loss for models answering two hop questions questions \emph{with} chain of thought with held out components. Values equal to or less than 0 indicate no generalization. For all components but Entity 1, at least one model generalizes, and for all components but Entity 2-Attribute Paris, at least one model does not generalize.}
    \label{fig:random_guessing_generalization_cot}
\end{figure}

\begin{minipage}{\textwidth}
\section{Additional Information Content Scaling Plots}
\begin{figure}[H]
    \centering
    \includegraphics[width=0.95\linewidth]{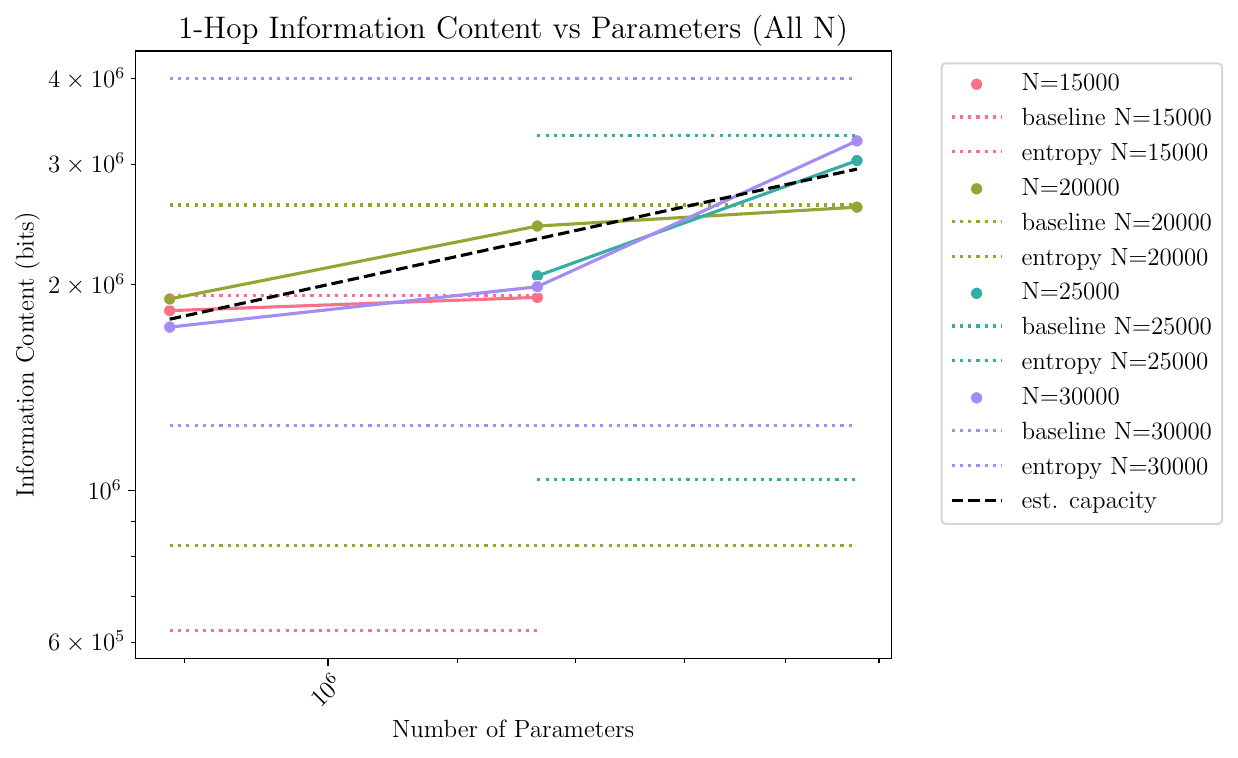}
    \caption{Additional measurements of 1 hop information content. These measurements were done using 4 relations instead of 17, and $\mu_p$ initialization was not used. The measured capacities were closer to 2 bits per parameter, but the trend less consistent than the main results.}
    \label{fig:1-hop-old}
\end{figure}
\end{minipage}

\begin{figure}[H]
    \centering
    \includegraphics[width=0.95\linewidth]{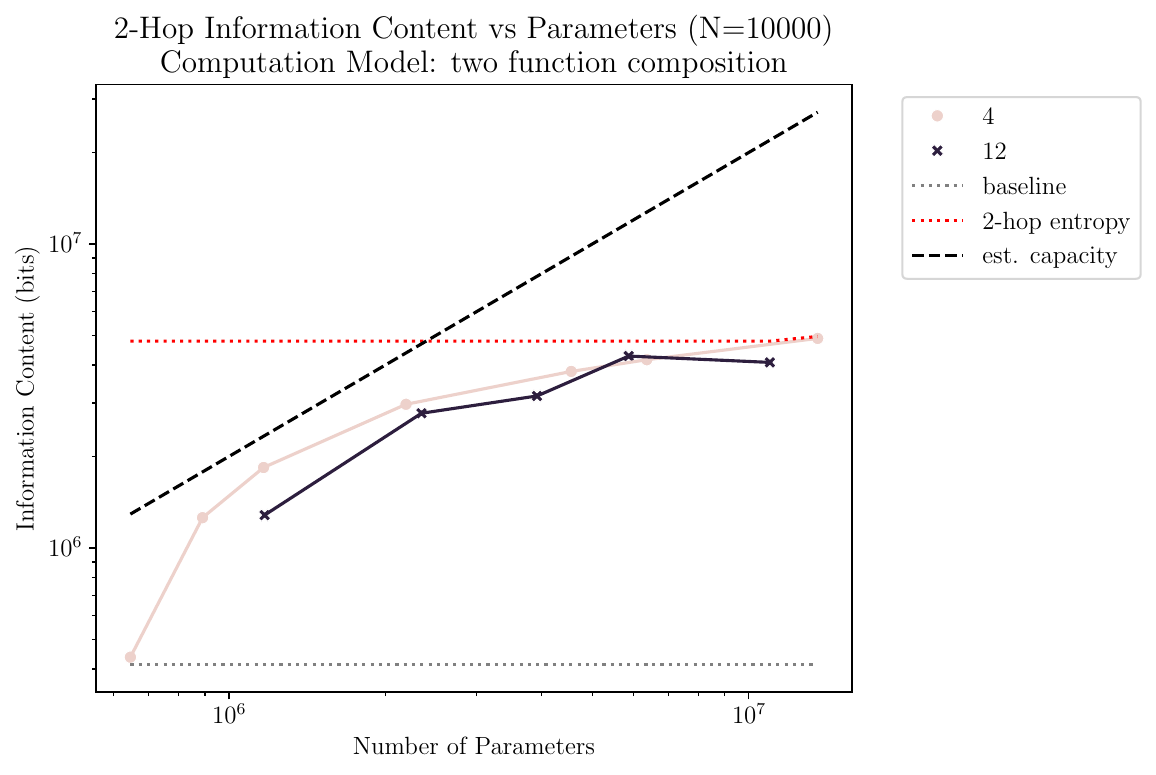}
    \caption{Comparison of information content of 4 layer and 12 layer models. The less consistent results of teh 12 layer models may be attributable to the fact that the hyperparameters were tuned on the 4 layer model and not readjusted for the deeper architecture.}
    \label{fig:2-hop-deep}
\end{figure}

\begin{minipage}{\textwidth}
\section{Probing results}\label{sec:probing}
\begin{table}[H]
\begin{flushleft}
\centering
\begin{tabular}{lccccc}
\hline
Model Details & Layer & Name Loss & Relation Loss & Attribute Loss & All Loss \\
\hline
Parameters: 890,320 & 1 & 5.58 & 6.67 & 6.70 & 5.56 \\
Relations: 4 & 2 & 5.56 & 6.61 & 6.66 & 5.26 \\
QA Loss: 3.16 & 3 & 5.62 & 6.53 & 6.68 & 5.14 \\
Inferred Algorithm: Memorization & 4 & 5.20 & 5.75 & 6.81 & 4.64 \\
\hline
Parameters: 890,320 & 1 & 5.57 & 6.38 & 6.40 & 5.57 \\
Relations: 17 & 2 & 5.84 & 6.21 & 6.40 & 5.65 \\
QA Loss: 8.18 & 3 & 5.94 & 6.22 & 6.40 & 5.76 \\
Inferred Algorithm: 2FC & 4 & 5.68 & 5.92 & 6.65 & 5.46 \\
\hline
Parameters: 2,192,400 & 1 & 5.26 & 6.39 & 6.40 & 5.27 \\
Relations: 17 & 2 & 5.60 & 6.10 & 6.40 & 5.28 \\
QA Loss: 6.56 & 3 & 5.69 & 6.09 & 6.40 & 5.34 \\
Inferred Algorithm: 2FC & 4 & 5.20 & 5.20 & 6.55 & 4.57 \\
\hline
Parameters: 6,369,424 & 1 & 4.57 & 6.38 & 6.40 & 4.57 \\
Relations: 17 & 2 & 4.92 & 5.98 & 6.23 & 4.46 \\
QA Loss: 2.25 & 3 & 5.02 & 6.01 & 6.26 & 4.59 \\
Inferred Algorithm: 2FC & 4 & 5.05 & 4.74 & 6.84 & 4.28 \\
\hline
Parameters: 13,575,328 & 1 & 3.98 & 6.38 & 6.40 & 3.99 \\
Relations: 17 & 2 & 4.47 & 5.47 & 5.98 & 3.67 \\
QA Loss: 0.19 & 3 & 4.69 & 5.59 & 6.09 & 3.92 \\
Inferred Algorithm: 2FC & 4 & 5.89 & 3.81 & 6.93 & 4.32 \\
\hline
\end{tabular}
\caption{Supervised probe losses for the second entity in two hop questions. An uninformed predictor achieves a loss of 6.47 (this is lower than QA loss as it was only evaluated on the first token of the relevant name). 2FC: two-function composition. Losses refer to probe positions: name loss is the loss of a probe trained on the (initial) name tokens, relation loss for a probe trained on relation tokens, attribute loss for attribute tokens and all loss for a probe trained on all previously mentioned tokens.}
\label{tab:loss_metrics}
\end{flushleft}
\end{table}
\end{minipage}

\begin{table}[H]
\begin{flushleft}
\centering
\begin{tabular}{lccccc}
\hline
Model Details & Function & Name Loss & Relation Loss & Attribute Loss & All Loss \\
\hline
Parameters: 890,320 & 1 & 5.89 & 6.64 & 6.67 & 5.63 \\
Relations: 4 & 2 & 5.97 & 6.62 & 6.64 & 5.56 \\
QA Loss: 3.16 & 3 & 5.80 & 6.59 & 6.63 & 5.71 \\
Algorithm: Memorization & 4 & 5.36 & 5.91 & 6.84 & 4.89 \\
\hline
Parameters: 890,320 & 1 & 5.61 & 6.44 & 6.46 & 5.58 \\
Relations: 17 & 2 & 5.88 & 6.26 & 6.46 & 5.69 \\
QA Loss: 8.18 & 3 & 5.98 & 6.27 & 6.45 & 5.79 \\
Algorithm: 2FC & 4 & 5.73 & 5.94 & 6.70 & 5.49 \\
\hline
Parameters: 2,192,400 & 1 & 5.31 & 6.44 & 6.47 & 5.30 \\
Relations: 17 & 2 & 5.65 & 6.15 & 6.40 & 5.32 \\
QA Loss: 6.56 & 3 & 5.74 & 6.15 & 6.39 & 5.37 \\
Algorithm: 2FC & 4 & 5.19 & 5.26 & 6.58 & 4.45 \\
\hline
Parameters: 6,369,424 & 1 & 4.56 & 6.43 & 6.46 & 4.55 \\
Relations: 17 & 2 & 4.94 & 6.03 & 6.30 & 4.46 \\
QA Loss: 2.25 & 3 & 5.05 & 6.06 & 6.34 & 4.63 \\
Algorithm: 2FC & 4 & 5.07 & 4.70 & 7.05 & 4.24 \\
\hline
Parameters: 13,575,328 & 1 & 4.07 & 6.44 & 6.46 & 4.04 \\
Relations: 17 & 2 & 4.58 & 5.60 & 6.10 & 3.76 \\
QA Loss: 0.19 & 3 & 4.82 & 5.72 & 6.21 & 4.04 \\
Algorithm: 2FC & 4 & 5.81 & 3.82 & 7.00 & 4.29 \\
\hline
\end{tabular}
\caption{Supervised probe losses for \emph{any} relation of the first entity in two hop questions. An uninformed predictor achieves a loss of 6.47. 2FC: two-function composition. Losses refer to probe positions: name loss is the loss of a probe trained on the (initial) name tokens, relation loss for a probe trained on relation tokens, attribute loss for attribute tokens and all loss for a probe trained on all previously mentioned tokens.}
\label{tab:loss_metrics_alt}
\end{flushleft}
\end{table}

\end{document}